\documentclass[twoside,leqno,twocolumn]{article}

\usepackage[letterpaper]{geometry}

\usepackage{ltexpprt}
\usepackage{hyperref}
\usepackage{amsmath}
\usepackage{autobreak}
\usepackage{multirow}
\usepackage{graphicx}
\usepackage{svg}
\usepackage{tikz,pgfplots}
    \usepgfplotslibrary{groupplots}
\usepackage{algorithm2e}

\pgfplotsset{compat=1.17}
\begin{document}

\newcommand\relatedversion{}

\title{\Large Citation Trajectory Prediction via Publication Influence Representation Using Temporal Knowledge Graph\relatedversion}

\author{Chang Zong\thanks{Zhejiang University, Hangzhou. zongchang@zju.edu.cn, yzhuang@zju.edu.cn, luwm@zju.edu.cn, jshao@zju.edu.cn, siliang@zju.edu.cn} \and Yueting Zhuang$^*$\thanks{Corresponding author.} \and Weiming Lu$^*$ \and Jian Shao$^*$ \and  Siliang Tang$^*$}

\date{}

\maketitle

\begin{abstract} \small\baselineskip=9pt Predicting the impact of publications in science and technology has become an important research area, which is useful in various real world scenarios such as technology investment, research direction selection, and technology policymaking. Citation trajectory prediction is one of the most popular tasks in this area. Existing approaches mainly rely on mining temporal and graph data from academic articles. Some recent methods are capable of handling cold-start prediction by aggregating metadata features of new publications. However, the implicit factors causing citations and the richer information from handling temporal and attribute features still need to be explored. In this paper, we propose CTPIR, a new citation trajectory prediction framework that is able to represent the influence (the momentum of citation) of either new or existing publications using the history information of all their attributes. Our framework is composed of three modules: difference-preserved graph embedding, fine-grained influence representation, and learning-based trajectory calculation. To test the effectiveness of our framework in more situations, we collect and construct a new temporal knowledge graph dataset from the real world, named AIPatent, which stems from global patents in the field of artificial intelligence. Experiments are conducted on both the APS academic dataset and our contributed AIPatent dataset. The results demonstrate the strengths of our approach in the citation trajectory prediction task.\end{abstract}

\section{Introduction}
Distinguishing high-impact publications is crucial to making decisions in business and research activities, such as investment in technology fields, selection of research topics, and development policymaking. Citations of a publication are usually applied to evaluate its potential impact. The question of how to predict citations has attracted more attention in recent years. With the development of knowledge graph technologies, the key issue is how to utilize the information provided by a knowledge graph to predict a publication's future citation trend.

Existing methods on this problem can be summarized in three ways. The first approach \cite{zhu2018citation, xiao2016modeling} tries to make use of prior knowledge and network techniques by assuming that citation trajectories obey the Power Law or log-normal functions. Traditional statistical methods are applied to make predictions. Another way \cite{li2019neural, abrishami2019predicting} focuses on taking advantage of text features of abstracts and reviews. Features are fed into recurrent neural network (RNN) models for time-series predictions. With the increasing popularity of graph neural networks (GNNs), recent works attempt to apply various structural learning models to exploit information from attributes of publications \cite{cummings2020structured, yu2012citation, jiang2021hints, li2021kernel, holm2020longitudinal}.

\begin{figure}[htbp]
    \centering
    \includegraphics[width=5cm]{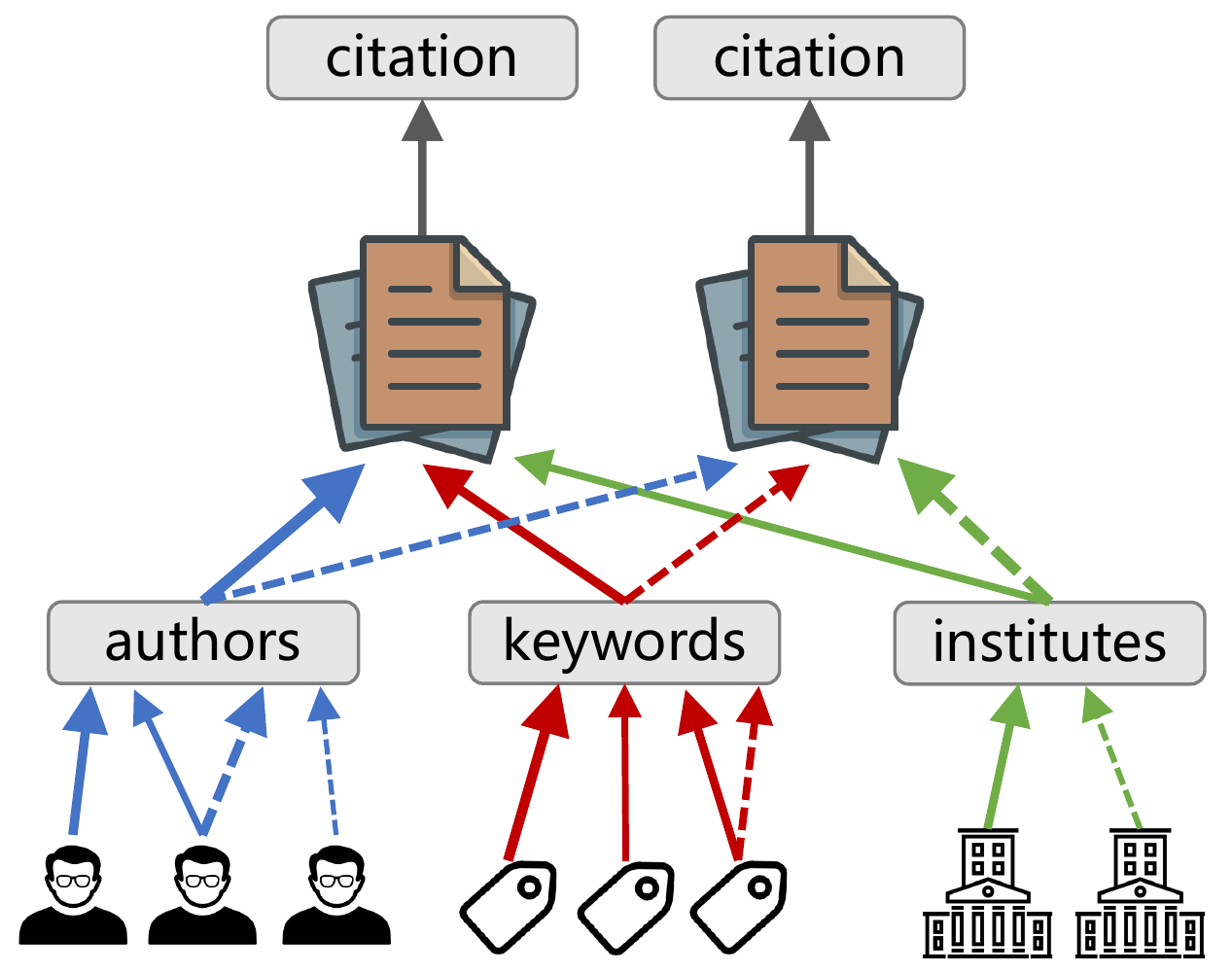}
    \caption{A diagram illustrating that citations of publications can be affected by their attributes and relation types in different levels.}
    \label{intro}
\end{figure}

However, citations are affected by many potential factors, and there is a lot of implicit information that must be considered in practice. For example, attributes of a publication, such as authors and keywords, should be treated significantly. The reputation of a scholar and the popularity of a field can greatly affect future citations of a publication. In addition, each attribute contributes to a publication at different levels (Figure \ref{intro}). The approaches in previous works simply apply GNNs to aggregate attribute features, which leads to a lack of fine-grained influence expression. On the basis of the above knowledge, a more powerful framework for predicting citation trajectories with the influence of publications derived from a temporal knowledge graph is needed to handle the problem: \textbf{How to represent and calculate the influence of a publication using as much of its information as possible?}

Current studies on temporal knowledge graphs try to manage changes in two adjacent snapshots, assuming that nodes should update smoothly or evolve dramatically \cite{gracious2021neural, trivedi2017know}. However, these assumptions require one to manually set a change rate to limit the evolution, which is not flexible. Existing works still mainly focus on handling structural and temporal features in separate steps, which leads to a lack of expression to treat dynamic graphs as a whole. Furthermore, accumulative citations are usually modeled as log-normal or cumulative distribution functions \cite{jiang2021hints, bai2019predicting}. It is still worth trying some alternatives to perform a further analysis. The potential enhancement mentioned above should be studied to handle another problem: \textbf{How can we improve the expressiveness of the framework for prediction tasks using temporal knowledge graphs?}

With the observations above, we propose CTPIR (\underline{C}itation \underline{T}rajectory \underline{P}rediction via \underline{I}nfluence \underline{R}epresentation), a new framework to predict citation trajectories with influence representation using temporal knowledge graphs. First, we optimize the R-GCN mechanism \cite{schlichtkrull2018modeling} to automatically learn the gaps between two adjacent snapshots. Second, we implement a fine-grained influence (citation momentum) representation module to make use of all historical information from a publication's attributes. Third, a learnable general logistic function is applied to fit the trajectories using the influence representation from the previous module.

We experiment our framework with two real world datasets. One is APS\footnote{https://journals.aps.org/datasets}, a public dataset of academic papers. Another, named AIPatent, is a new dataset that we construct with global patents in the field of artificial intelligence. Compared to some baselines, the results show that CTPIR outperforms those methods in all cases.

Our key contributions are summarized in the following points:
\begin{itemize}
\item \textbf{Novel framework}: We propose a new framework, named CTPIR, which implements a fine-grained influence representation approach using a more expressive temporal graph learning process and optimizes existing methods to bring prediction results much closer to observations.
\item \textbf{Improved evaluation}: We construct a new temporal knowledge graph dataset named AIPatent for the task, which is also a strong supplement for the community to carry out various temporal graph studies. We also design and implement multiple subtasks to evaluate approaches from a more comprehensive view.
\item \textbf{Multifaceted analysis}: We analyze the experimental performance from multiple aspects. Explanations on how CTPIR performs better compared to other recent approaches are discussed. Some weaknesses and further efforts are also mentioned to guide future studies.
\end{itemize}
The dataset we use in this work, including our AIPatent contributed dataset, and the code to reproduce are available in our GitHub repository: 
\href{https://github.com/changzong/CTPIR}{https://github.com/changzong/CTPIR}

\section{Related Work}
\subsection{Citation and Popularity Prediction.}
Modern approaches to citation count prediction (CCP) aim to combine attribute information with temporal features. GNNs are commonly used to capture topological features of citation networks. The encoded nodes are sent to RNNs or attention models for time-series forecasting. A previous work \cite{holm2020longitudinal} follows this simple encoder-decoder architecture. Some previous studies \cite{xu2020casgcn, li2021kernel, tang2021fully, zhao2022utilizing} put emphasis on cascade graphs for popularity prediction, using a kernel method to estimate structural similarities. These works are based on simply combining graph embedding with time-series methods. In contrast, we introduce a method to fully utilize all past characteristics of a publication's attributes. A recent work called HINTS \cite{jiang2021hints} adds an imputation module to aggregate the information from each snapshot of graphs. Another work proposes a heterogeneous dynamical graph neural network (HDGNN) \cite{zhou2020heterogeneous} to predict the cumulative impact of articles and authors. The latest work \cite{huang2022fine} uses an attention mechanism to represent the sequence of content from citation relations. Although these works can take advantage of richer information, their lack of fine-grained design to represent the influence of a publication is not conducive to achieving good prediction performance.

\subsection{Temporal Graph Embedding.}
We focus on deep learning-based temporal graph embedding approaches. Several previous works implement a straightforward way to combine GCN and RNN models to extract structural and temporal features \cite{park2019exploiting,shrestha2019learning,cai2021structural}. RNN variants are applied as the temporal module to perform downstream tasks such as anomaly detection. Meanwhile, temporal attention models can be a substitute for GCN to extract topological features \cite{xu2019spatio, li2019learning,park2020st}. A recent paper \cite{beladev2020tdgraphembed} tries to represent global structural information of graphs at each timestamp, rather than focus solely on nodes. Another article \cite{leblay2020towards} concerns the granularity of the timestamp and attempts to take into account the precision of arbitrary times. Tensor decomposition is also a useful approach to represent evolution patterns in graphs \cite{ma2021temporal}. The work \cite{you2022roland} further extends the architecture of a static GNN to dynamic graphs by updating hierarchical node embeddings. Some modern approaches \cite{gao2022novel, zhong2022dynamic} apply contrast learning or the transformer model to the traditional GCN framework for additional information. Instead of extracting temporal features after encoding structural information, we find that applying the RNN method on the history information of each attribute first can utilize more previous features for downstream tasks.

\section{Problem Statement}
\begin{figure*}[htbp]
    \centering
    \includegraphics[width=17cm]{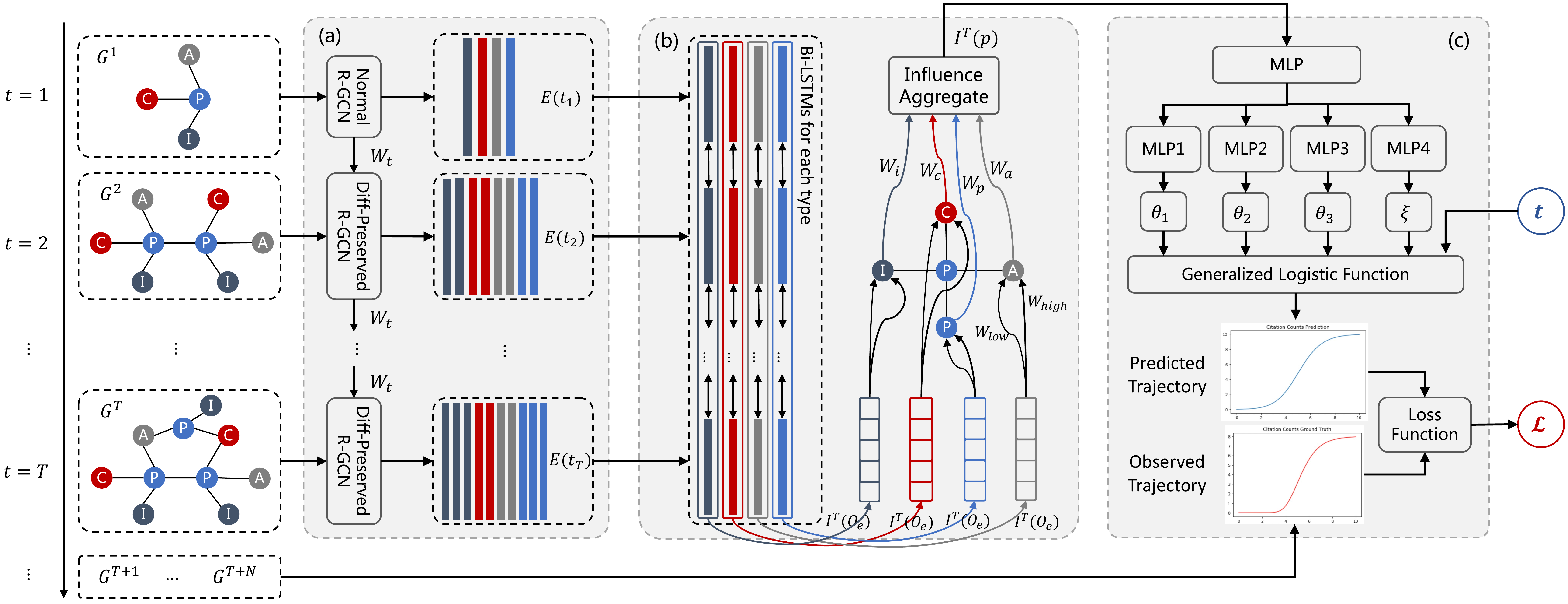}
    \caption{Framework of CTPIR, which takes previous graph snapshots as input, and successively executes through three steps: (a) difference-preserved graph embedding, (b) fine-grained influence representation, (c) learning-based trajectory calculation, and finally gets a learnable loss as output.}
    \label{CTPIR}
\end{figure*}
\subsection{Preliminaries.}
Following \cite{cai2022temporal}, we define a knowledge graph that contains timestamps as $G=(E,R,A,T,F)$, where $E$, $R$, $A$, $T$, and $F$ are sets of entities, relations, attributes, timestamps, and facts. We denote a temporal knowledge graph as a sequence of snapshots over time, denoted by $G_{tmp}^T=\lbrace G^1,G^2,...,G^T \rbrace$, where $G^t=(E^t,R^t,A^t,F^t)$ $(1 \le t \le T)$ is a snapshot with its entities and relations at the time $t$.

\subsection{Problem Definition.}
For a publication $p$ in the year of $T$, the sequence of citation counts of $p$ in the next $N$ years can be noted as $\hat{C_p}=\lbrace C_p^{T+1}, C_p^{T+2}, ... , C_p^{T+N} \rbrace$, where $C_p^{T+N}$ is the citation count in the year $T+N$. Then, Given a temporal knowledge graph $G_{tmp}^T$ with snapshots from year $1$ to $T$, our goal is to learn a function $f(\cdot)$, which maps the publication $p$ to its sequence of citation counts $\hat{C_p}$ in the next $N$ years. The problem is described in the following.
\begin{small}
\begin{equation}
\begin{aligned}
\hat{C_p} &= f(G_{tmp}^T, p) \\
G_{tmp}^T &= \lbrace G^1,G^2,...,G^T \rbrace\\ 
\hat{C_p} &= \lbrace C_p^{T+1}, C_p^{T+2}, ... , C_p^{T+N} \rbrace
\label{eq:eq1}
\end{aligned}
\end{equation}
\end{small}

\section{Proposed Framework: CTPIR}
We now introduce our proposed CTPIR. The framework is designed with three motivations as follows:
\begin{itemize}
\item Evolutionary differences between two adjacent snapshots should be preserved dynamically, as different networks may have entirely different behaviors \cite{kashtan2007varying}. 
\item Citations of a publication can be affected by all historical information on its attributes, which can be denoted as \textbf{influence}. Our approach should quantify the influence of all entities and their different contributions to a publication for feature aggregation.
\item The growth of citations can be viewed as the prevalence of publications. The general logistic function, a widely used function to model disease prevalence \cite{lee2020estimation}, may fit well with the citation trajectories.
\end{itemize}

\subsection{Difference-preserved Graph Embedding.}
Given a temporal knowledge graph generated from publications. Our framework first realizes an embedding method to extract node features from these graph snapshots. We employ a relational graph convolution network (R-GCN) \cite{schlichtkrull2018modeling} and modify it to preserve the evolutionary difference for the same nodes between adjacent snapshots, using a learnable parameter. The hidden layer embeddings can be updated as: 
\begin{small}
\begin{equation}
\begin{split}
h_{i,t}^{l+1} = \sigma(\sum_{r\in R}\sum_{j\in N^r_{i,t}}W_r^{(l)}h_{j,t}^{(l)}+W_0^{(l)}h_{i,t}^{(l)}&+\\\sum_{r\in R}\sum_{k\in N^r_{i,t-1}}W_t^{(l)}W_r^{(l)}h_{k,t-1}^{(l)}&+\\W_t^{(l)}W_0^{(l)}h_{i,t-1}^{(l)}) \ ,
\end{split}
\label{eq:eq2}
\end{equation}
\end{small} where $h_{i,t}^{(l+1)}$ denotes the feature vector of the node $i$ in layer $l + 1$ at the time $t$, $r$ represents a relation type in the set $R$, $N^r_{i,t}$ is the set of neighbors of the node $i$ with the relation type $r$ at the time $t$, $W_0^{(l)}$ is the aggregation weight for the node $i$ in layer $l$. $W_r^{(l)}$ is the aggregation weight for neighbors with relation $r$ in layer $l$, $W_t^{(l)}$ is a feature transformation weight from the time $t-1$ to the time $t$, $\sigma$ is an activation function. For $t=0$, as there is no previous adjacent feature can be transformed, the model is degenerated to a normal R-GCN layer.

\subsection{Fine-grained Influence Representation.}
In this module, we first try to quantify the influence of each attribute of a publication using sequential past features generated from the previous module.

We apply a Bi-LSTM model followed by a fully connected layer to digest the sequence of feature vectors for each attribute. We set an independent Bi-LSTM model for each relation type to process the related attribute entities. The equation to represent the influence of an attribute related to a publication is defined as follows:  
\begin{small}
\begin{equation}
\begin{aligned}
I^{T}(O_e(p,r)) &= FC_r(\mathop{LSTM_r} \limits ^{\rightharpoonup}(seq)||\mathop{LSTM_r} \limits ^{\leftharpoonup}(seq)) \\
seq &= \lbrace V_e^1(p,r), V_e^2(p,r), ... V_e^T(p,r) \rbrace \ ,
\label{eq:eq3}
\end{aligned}
\end{equation}
\end{small} where $I^{T}(\cdot)$ is the value of influence at the time $T$, $O_e^r(p,r)$ is the object of attribute entity $e$ related to the publication $p$ with the relation type $r$, $FC_r(\cdot)$ is a fully connected operation for the type $r$, $\mathop{LSTM_r} \limits ^{\rightharpoonup}(seq)$ and $\mathop{LSTM_r} \limits ^{\leftharpoonup}(seq)$ represents LSTM layers of two directions, the notation $||$ means a concatenate operation, $seq$ is the past feature sequence of an entity, $V_e^T(p,r)$ is the feature vector of entity $e$ at the time $T$ output from the last layer using Equation\ref{eq:eq2}. 

The influence of attributes should be aggregated to represent the overall influence of a publication, which is treated as a momentum to be cited in the future. We assume that attributes can affect publications to different levels, considering their positions and types. For example, the first author of a publication plays a more significant role than the others. We defined two parameters $W_{high}$ and $W_{low}$ as proxies of higher-level and lower-level effects, respectively. Furthermore, different relationships may affect a publication to different degrees. We then set another parameter $W_r$ to represent the contribution according to each relationship. The influence of a publication is computed as follows:  
\begin{small}
\begin{equation}
\begin{split}
I^{T}(p) = \sum_{r\in R} W_r \cdot \sum_{e \in E_r}(I^T(O_e(p,r)) \cdot W_{high} &+\\ I^T(O_e(p,r) \cdot W_{low}) \ ,
\end{split}
\label{eq:eq4}
\end{equation}
\end{small} where $I^T(p)$ represents the overall influence of the publication $p$ at the time $T$, $I^T(O_e(p,r))$ is the influence of $p$'s attribute from Equation\ref{eq:eq3}, $R$ is the set of relation types and $E_r$ is the set of entities with relation type $r$. $W_{high}$ and $W_{low}$ are the higher-level and lower-level effects, $W_r$ is the relational contribution to the publication $p$.

\subsection{Learning-based Trajectory Calculation.}
Treating citations as the prevalence of publications, we employ a generalized logistic function widely used in pandemic prevalence \cite{lee2020estimation} to fit citation trajectories. All four parameters of the function can be learned and updated from the framework by learning four distinct MLP (multilayer perceptron) models. Finally, given the influence representation generated from the previous module, the future sequence of cumulative citation counts can be predicted using the following: 
\begin{small}
\begin{equation}
\begin{split}
C &= \frac{\theta_1(\mathcal{M})}{1+\xi(\mathcal{M})exp[-\theta_2(\mathcal{M}) \cdot (t-\theta_3(\mathcal{M}))]^{1/{\xi(\mathcal{M})}}} \\ 
\mathcal{M} &= I^T(p) \ ,
\end{split}
\label{eq:eq5}
\end{equation}
\end{small} where $\mathcal{M}$ is the momentum of citation, which is equal to the influence representation of the publication $p$ at the time $T$, $C(\mathcal{M}, t)$ donates the mapping function to calculate the citation count at the time $t$ using the influence of $p$, $\theta_1(\mathcal{M})$ can be treated as the peak of citation counts, $\theta_2(\mathcal{M})$ shows the rising rate of being cited, $\theta_3(\mathcal{M})$ can denote the lag phase before the publication is firstly cited, and $\xi(\mathcal{M})$ indicates the smoothness of the curve.

\subsection{Loss Function.}
Following previous studies, we apply MALE (mean absolute logarithmic error) and RMSLE (square root mean logarithmic error) as our loss functions to evaluate our CTPIR framework and baselines. They are log-scaled versions of MAE and RMSE, which are commonly used in prediction tasks but are sensitive to outliers. The loss is the average value for all prediction years and target publications. Two functions are presented in the following.
\begin{small}
\begin{equation}
\begin{split}
L_{MALE} &= \frac{1}{M}\sum ^{M-1}_{j=0}\frac{1}{N}\sum ^{N-1}_{i=0} \lvert {log{\hat{C}_{i,j}}-log{C_{i,j}}} \rvert \\
L_{RMSLE} &= \sqrt{\frac{1}{M}\sum^{M-1}_{j=0}\frac{1}{N}\sum^{N-1}_{i=0}(log{\hat{C}_{i,j}}-log{C_{i,j}})^2} \ ,
\end{split}
\label{loss} 
\end{equation}
\end{small} where $\hat{C}_{i,j}$ is our predicted citation counts for the $j_{th}$ publication in the $i_{th}$ year, $C_{i,j}$ is the corresponding observed citation counts, $M$ is the total number of publications, $N$ is the number of years we desire to predict in the future.

\section{AIPatent Dataset}
We provide a new temporal knowledge graph dataset named \textbf{AIPatent}. We notice that the quality and industry relevance of a dataset is important when studying temporal knowledge graphs in real situations. Existing public datasets usually suffer from inaccuracy and are only used for academic scenarios. AIPatent is collected and constructed from a global patent commercial platform filtered by tags related to artificial intelligence. The dataset is already uploaded and available from our GitHub repository.

\begin{figure}[htbp]
    \centering
    \includegraphics[width=8.5cm]{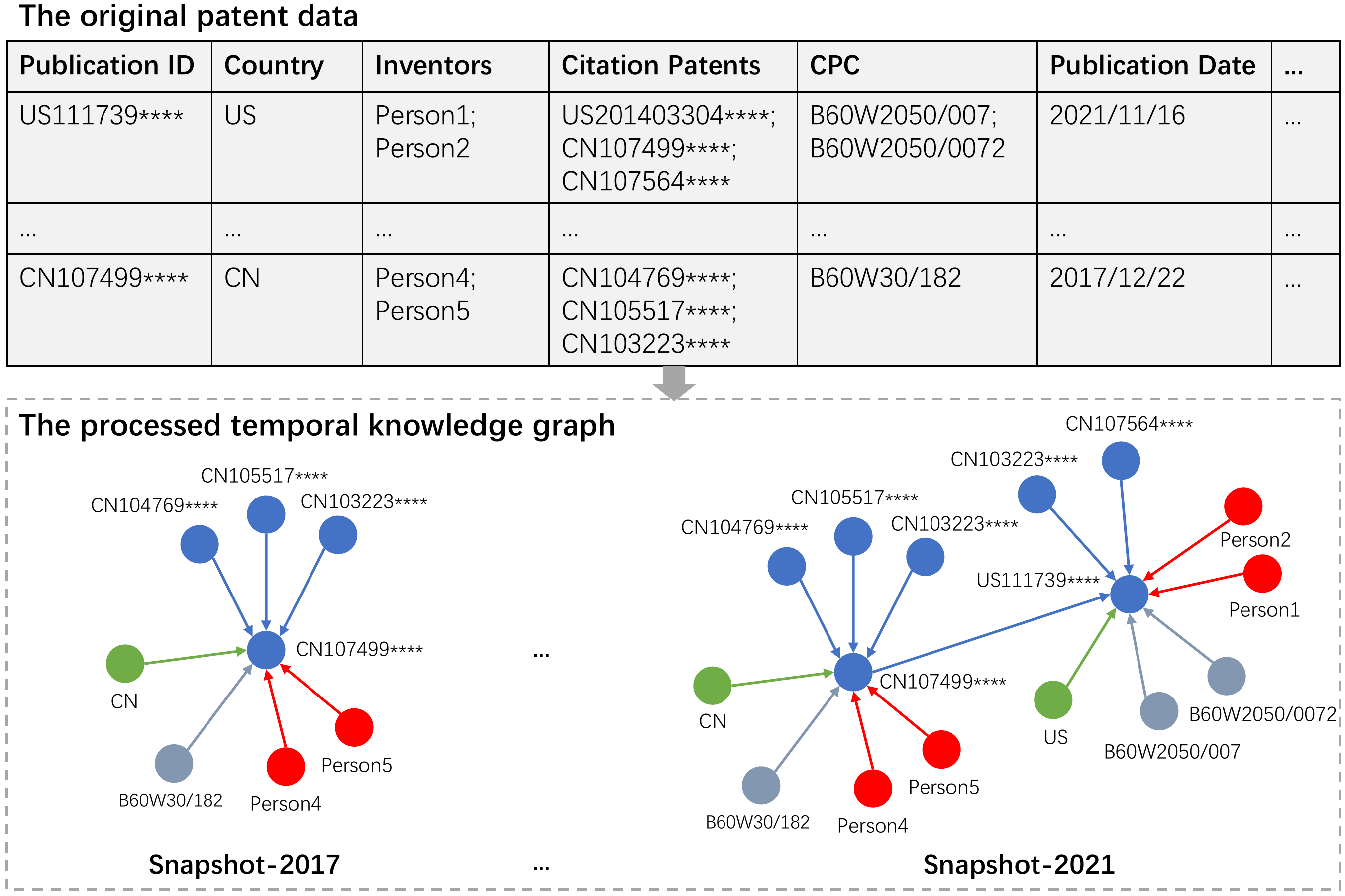}
    \caption{A snippet of dataset that we use in CTPIR. Some data is obfuscated. \textbf{Above} is the original data we collect. \textbf{Below} is the processed temporal knowledge graph.}
    \label{snippet}
\end{figure}

\begin{table*}\small
\centering
\setlength{\belowcaptionskip}{0.4cm}
 \caption{Dataset statistics of a segment of AIPatent from the year 2015 to 2021.}
 \begin{tabular}{c|c|ccccccc}
   \textbf{Relation Types} & \textbf{Properties} &\textbf{2015}&\textbf{2016}&\textbf{2017}&\textbf{2018}&\textbf{2019}&\textbf{2020}&\textbf{2021}\\
  \hline 
  \multirow{3}{*}{\textbf{citedBy}}&$\vert E \vert$ &156684&197290&254438&332312&430199&521963&607809\\
  ~&$\vert R \vert$&560958&715101&937972&1211852&1592095&1993470&2371361\\
  ~&$\bar D$&7.1603&7.2462&7.3729&7.2934&7.4016&7.6383&7.8030\\
  \hline 
  \multirow{3}{*}{\textbf{relatedTo}}&$\vert E \vert$&346023&419615&521160&654882&830763&1044713&1245627\\
  ~&$\vert R \vert$&2448072&3155158&4150369&5469801&7224994&9322740&11244536\\
  ~&$\bar D$&14.1498&15.0382&15.9273&16.7046&17.3936&17.8474&18.0543\\
  \hline 
  \multirow{3}{*}{\textbf{appliedBy}}&$\vert E \vert$&476775&559548&672545&820147&1011069&1245065&1465285\\
  ~&$\vert R \vert$&510239&591512&699531&840231&1024256&1252572&1469457\\
  ~&$\bar D$&2.1404&2.1267&2.1142&2.0802&2.0490&2.0121&2.0057\\
  \hline 
  \multirow{3}{*}{\textbf{belongTo}}&$\vert E \vert$&309520&378852&475442&603826&772219&980661&1178524\\
  ~&$\vert R \vert$&309451&378781&475371&603754&772147&980589&1178452\\
  ~&$\bar D$&1.9995&1.9996&1.9997&1.9998&1.9998&1.9998&2.0000\\
\end{tabular}
 \label{statistics}
\end{table*}

\subsection{Dataset Construction.}
For extracting patents related to fields of artificial intelligence, we first collect CPC codes (Cooperative Patent Classification) that refer to the PATENTSCOPE Artificial Intelligence Index \footnote{https://www.wipo.int/tech\_trends/en/artificial\_intelligence/
patentscope.html}. Then, patents published between the year 2002 and 2021 are filtered using these CPC codes from the platform. Patents are downloaded and processed to generate our temporal knowledge graphs with Python scripts. We finally get 20 heterogeneous snapshots divided by year, and each one is saved in an individual file along with randomly initialized feature vectors for each node. A snippet of the original data and the processed snapshots is shown in Figure \ref{snippet}.

\subsection{Dataset Analysis.}
\begin{figure}[htbp]
    \centering
    \includegraphics[width=8.5cm]{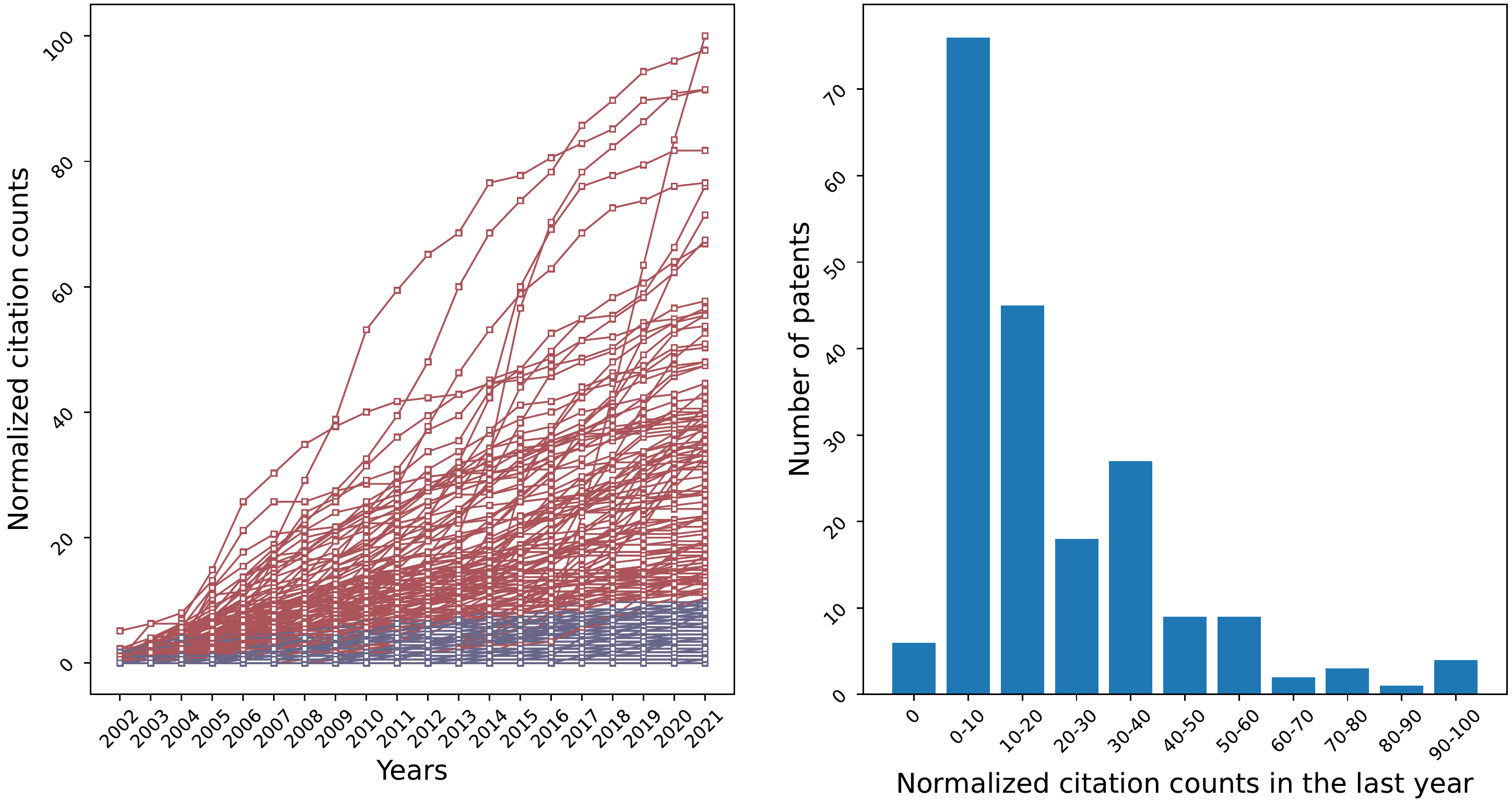}
    \caption{Citation statistics in AIPatent. \textbf{Left}: Citation trajectories of 200 randomly selected patents over 20 years. Red lines reflect citations above the average. Purple lines are for citations below the average. \textbf{Right}: The distribution of accumulative citation counts in the last year.}
    \label{dataset}
\end{figure}

Each snapshot in AIPatent is divided into four subgraphs according to the relationships. Some network analysis is performed, and three graph properties are shown in Table\ref{statistics}, including the number of entities indicated as $\vert E \vert$, the number of relations indicated as $\vert R \vert$, and the average degree indicated as $\bar{D}$. Furthermore, Figure \ref{dataset} shows the citation trajectories and the citation distribution of some randomly selected nodes in AIPatent. It tells us that the more citations accumulate, the fewer patents we can observe. We notice that the average degrees shown in the $citedBy$ subgraph are around 7, which is close to the number stated by the existing citation networks.

\section{Experiments}
\subsection{Experimental Settings.}
\textit{(1) Datasets.} For AIPatent, we use the first ten snapshots for temporal graph embedding, and the rest for prediction and evaluation. The influence of patents in 2012 is generated to predict citation counts in the following 10 years. APS (American Physical Society) is a dataset of academic papers widely used for social science studies. We form a temporal knowledge graph using APS papers from 1995 to 2004. Ten successive graph snapshots are constructed. We used the first five snapshots for feature learning and the last five for evaluation. 10000 samples are randomly selected from two datasets, and two-thirds of them are used for training, the rest are for testing.

\textit{(2) Implementation details.} We implement CTPIR using PyTorch 1.7. In the difference-preserved graph embedding module, we set up a two-layer R-GCN model with an embedding size of 128 and a hidden size of 64. In the influence representation module, we set both the hidden size of Bi-LSTM and the output size to 128. In the trajectory prediction module, we set up four separate three-layer MLP models to learn the parameters of the generalized logistic function. In general, we set 512 for batch size and 0.05 for learning rate in all training processes. All experiments are carried out on a single desktop machine with an Nvidia A100 GPU. We run 20 epochs for each experiment, since we notice that the performance can hardly improve after it. We keep the best result of 5 attempts for each experiment.

\subsection{Performance Comparison.}

\textit{(1) Baselines.} We examine the performance of CTPIR compared to a variety of baselines. 

\begin{itemize}
\item \textbf{GRU+MLP} is a basic but widely used framework for time-series prediction problems. Without any graph embedding method, history citation counts are directly used for predicting following sequences.
\item \textbf{TGNN} \cite{holm2020longitudinal} uses a straightforward encoder-decoder framework for citation count prediction, which contains a GCN model for embedding and a LSTM model for sequence generation.
\item \textbf{HINTS} \cite{jiang2021hints} is another framework for predicting citation count consisting of three modules: graph encoder, weighted metadata imputation, and a citation curve generator. 
\item \textbf{HDGNN} \cite{zhou2020heterogeneous} is a more recent framework that can extract the feature of a publication by aggregating both attributes and neighboring nodes with RNN and attention models.
\item \textbf{DeepCCP} \cite{zhao2022utilizing} is another latest approach to model the citation network in a cascade graph and utilizing an end-to-end deep learning framework to encode both structural and time-series information.
\item \textbf{CTPIR+GCN} and \textbf{CTPIR+MLP} are two modified versions based on our CTPIR, by replacing the graph embedding and count prediction modules with simple GCN and MLP models, respectively, to verify our primary contribution.
\end{itemize}

\textit{(2) Subtask design.} Three subtasks are performed in our work. The first task, denoted \textbf{Newborn Task}, is to predict citation counts in new publications that are rarely cited, with citation counts less than 5 (for AIPatent) and 2 (for APS). The second task is to predict publications that have been cited many times in previous years, with citations greater than 30 (for APS) and 12 (for AIPatent), denoted \textbf{Grown Task}. The third task is for samples randomly derived from knowledge graphs without citation count limit, denoted \textbf{Mix Task}. These three subtasks are trained with the same hyperparameters.

\begin{table*}[htbp]\small
\centering
\setlength{\belowcaptionskip}{0.4cm}
\caption{Performance results using APS (\textbf{Above}) and AIPatent (\textbf{Below}) to predict citation counts. Experiments are performed on three subtasks with CTPIR and baselines.}
\label{performance1}
\begin{tabular}{|l|lll|lll|}
\hline
\multirow{2}{*}{\textbf{Frameworks}}  & \multicolumn{3}{c|}{\textbf{RMSLE}}  &\multicolumn{3}{c|}{\textbf{MALE}}
\\ \cline{2-7} & 
\multicolumn{1}{l|}{\textbf{Mix Task}} & \multicolumn{1}{l|}{\textbf{Newborn Task}} & \textbf{Grown Task} & \multicolumn{1}{l|}{\textbf{Mix Task}} & \multicolumn{1}{l|}{\textbf{Newborn Task}} & \textbf{Grown Task} 
\\ \hline      
\multirow{1}{*}{GRU+MLP} & \multicolumn{1}{l|}{0.9233} & \multicolumn{1}{l|}{0.5846} & 0.4755 & \multicolumn{1}{l|}{0.7893} & \multicolumn{1}{l|}{0.5024} & 0.3179
\\
\multirow{1}{*}{TGNN} & \multicolumn{1}{l|}{0.8964} & \multicolumn{1}{l|}{0.5446} & 0.4325 & \multicolumn{1}{l|}{0.7661} & \multicolumn{1}{l|}{0.4759} & 0.2972          
\\  
\multirow{1}{*}{HINTS} & \multicolumn{1}{l|}{0.8973} & \multicolumn{1}{l|}{0.5539} & 0.3588 & \multicolumn{1}{l|}{0.7723} & \multicolumn{1}{l|}{0.4901} & 0.2950
\\  
\multirow{1}{*}{DeepCCP} & \multicolumn{1}{l|}{0.9013} & \multicolumn{1}{l|}{0.5426} & 0.4342 & \multicolumn{1}{l|}{0.7661} & \multicolumn{1}{l|}{0.4747} & 0.2956
\\  
\multirow{1}{*}{HDGNN} & \multicolumn{1}{l|}{0.8988} & \multicolumn{1}{l|}{0.5419} & 0.3786 & \multicolumn{1}{l|}{0.7652} & \multicolumn{1}{l|}{0.4716} & 0.2471
\\ \hline 
\multirow{1}{*}{CTPIR+GCN} & \multicolumn{1}{l|}{0.6671} & \multicolumn{1}{l|}{0.2912} & 0.3234 & \multicolumn{1}{l|}{0.5133} & \multicolumn{1}{l|}{\textbf{0.1943}} & 0.2280
\\
\multirow{1}{*}{CTPIR+MLP} & \multicolumn{1}{l|}{0.7078} & \multicolumn{1}{l|}{0.3541} & 0.3296 & \multicolumn{1}{l|}{0.5506} & \multicolumn{1}{l|}{0.2702} & 0.2451
\\
\multirow{1}{*}{\textbf{CTPIR}} & \multicolumn{1}{l|}{\textbf{0.5864}} & \multicolumn{1}{l|}{\textbf{0.2883}} & \textbf{0.2641} & \multicolumn{1}{l|}{\textbf{0.4005}} & \multicolumn{1}{l|}{0.2050} & \textbf{0.1916}  
\\ \hline 
\end{tabular}
\begin{tabular}{|l|lll|lll|}
\hline
\multirow{2}{*}{\textbf{Frameworks}}  & \multicolumn{3}{c|}{\textbf{RMSLE}}  &\multicolumn{3}{c|}{\textbf{MALE}}
\\ \cline{2-7} & 
\multicolumn{1}{l|}{\textbf{Mix Task}} & \multicolumn{1}{l|}{\textbf{Newborn Task}} & \textbf{Grown Task} & \multicolumn{1}{l|}{\textbf{Mix Task}} & \multicolumn{1}{l|}{\textbf{Newborn Task}} & \textbf{Grown Task} 
\\ \hline      
\multirow{1}{*}{GRU+MLP} & \multicolumn{1}{l|}{0.9858} & \multicolumn{1}{l|}{0.5843} & 0.5261 & \multicolumn{1}{l|}{0.7901} & \multicolumn{1}{l|}{0.4385} & 0.5143
\\ 
\multirow{1}{*}{TGNN} & \multicolumn{1}{l|}{0.9644} & \multicolumn{1}{l|}{0.5647} & 0.5060 & \multicolumn{1}{l|}{0.7745} & \multicolumn{1}{l|}{0.4101} & 0.5029
\\ 
\multirow{1}{*}{HINTS} & \multicolumn{1}{l|}{0.9783} & \multicolumn{1}{l|}{0.5780} & 0.4826 & \multicolumn{1}{l|}{0.7852} & \multicolumn{1}{l|}{0.4379} & 0.4912
\\
\multirow{1}{*}{DeepCCP} & \multicolumn{1}{l|}{0.9685} & \multicolumn{1}{l|}{0.5695} & 0.5083 & \multicolumn{1}{l|}{0.7814} & \multicolumn{1}{l|}{0.4363} & 0.4028
\\  
\multirow{1}{*}{HDGNN} & \multicolumn{1}{l|}{ 0.9640} & \multicolumn{1}{l|}{0.5651} & 0.4989 & \multicolumn{1}{l|}{0.7734} & \multicolumn{1}{l|}{0.4171} & 0.3970
\\ \hline 
\multirow{1}{*}{CTPIR+GCN} & \multicolumn{1}{l|}{0.6941} & \multicolumn{1}{l|}{0.5564} & 0.3920 & \multicolumn{1}{l|}{0.5279} & \multicolumn{1}{l|}{0.4042} & 0.2946
\\
\multirow{1}{*}{CTPIR+MLP} & \multicolumn{1}{l|}{0.7195} & \multicolumn{1}{l|}{\textbf{0.5329}} & 0.5007 & \multicolumn{1}{l|}{0.5544} & \multicolumn{1}{l|}{\textbf{0.3741}} & 0.3833
\\
\multirow{1}{*}{\textbf{CTPIR}} & \multicolumn{1}{l|}{\textbf{0.5995}} & \multicolumn{1}{l|}{0.5338} & \textbf{0.3855} & \multicolumn{1}{l|}{\textbf{0.4156}} & \multicolumn{1}{l|}{0.3801} & \textbf{0.2787} 
\\ \hline
\end{tabular}
\end{table*}

\textit{(3) Results.} The results of the three subtasks are shown in Table\ref{performance1}. We observe that CTPIR can greatly reduce errors on all subtasks compared to baselines (33.27\%, 47.76\%, 36.44\% on APS and 38.14\%, 7.11\%, 24.01\% on AIPatent, demonstrating that our attempt to quantify the fine-grained influence of publications is encouraging. Considering most cases of Newborn Task, our CTPIR variants achieve the best in predicting newly published patents, showing that a simple sequence model is adequate when there is a lack of citation information.

\subsection{Ablation Study.}
\begin{figure}[htbp]
    \centering
    \includegraphics[width=8.5cm]{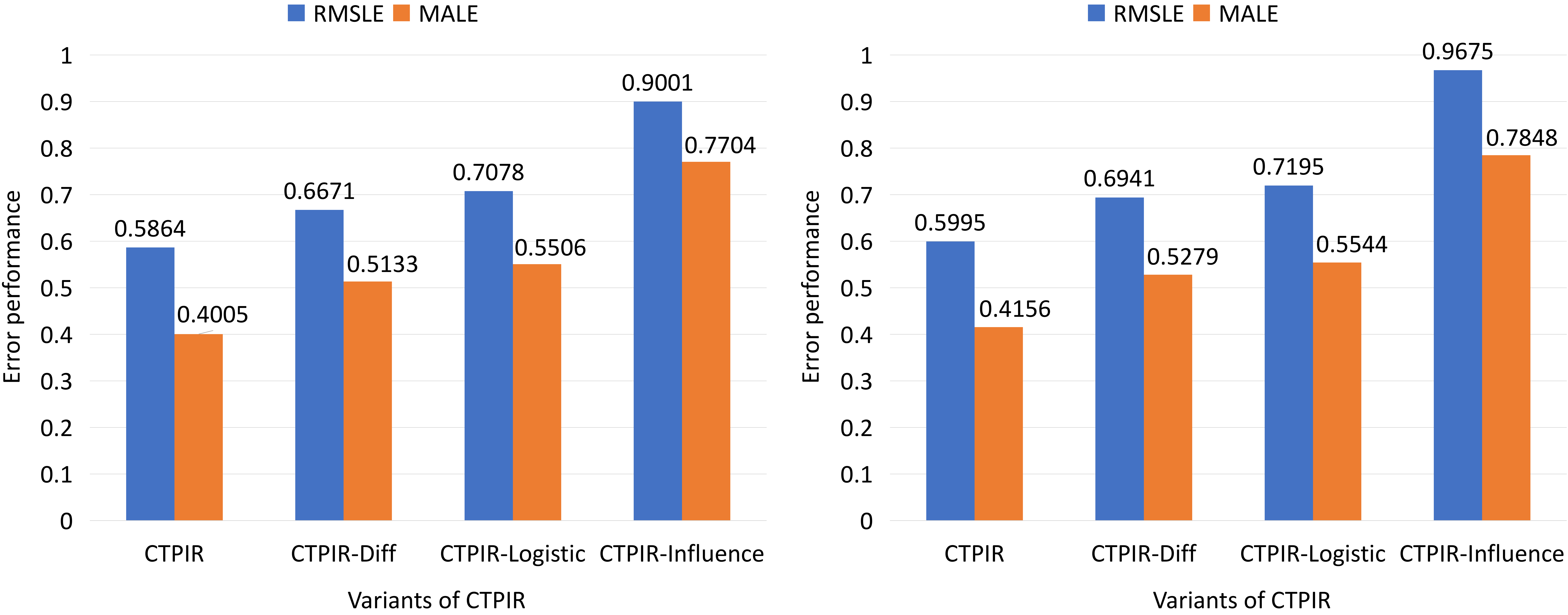}
    \caption{An ablation analysis with variants of CTPIR on APS (\textbf{Left}) and AIPatent (\textbf{Right}) datasets. The influence representation module is the primary factor driving the performance.}
    \label{ablation}
\end{figure}

We denote "CTPIR-X" as the variant of CTPIR to replace the module "X" by a similar or simpler one. The charts in Figure \ref{ablation} show the performance of these variants.

\textbf{CTPIR-Influence} is a framework that directly aggregates attribute features, rather than using all the history information from each entity. The tremendous degeneration in two datasets (54.51\% on APS and 61.38\% on AIPatent) demonstrates that the influence representation module dominates performance. In addition, a smaller reduction (13.74\% on APS and 15.78\% on AIPatent) occurs in \textbf{CTPIR-Diff}, which simply replaces our difference-preserved module with a normal R-GCN. Moreover, \textbf{CTPIR-Logistic} applies a commonly used log-normal function instead of our generalized logistic function. A decrease (20.70\% on APS and 20.02\% on AIPatent) shows that our hypothesis of treating citations as the prevalence of publications is close to real world situations.

\subsection{Time Distance Analysis.}
\begin{figure}[htbp]
    \centering
    \includegraphics[width=8.5cm]{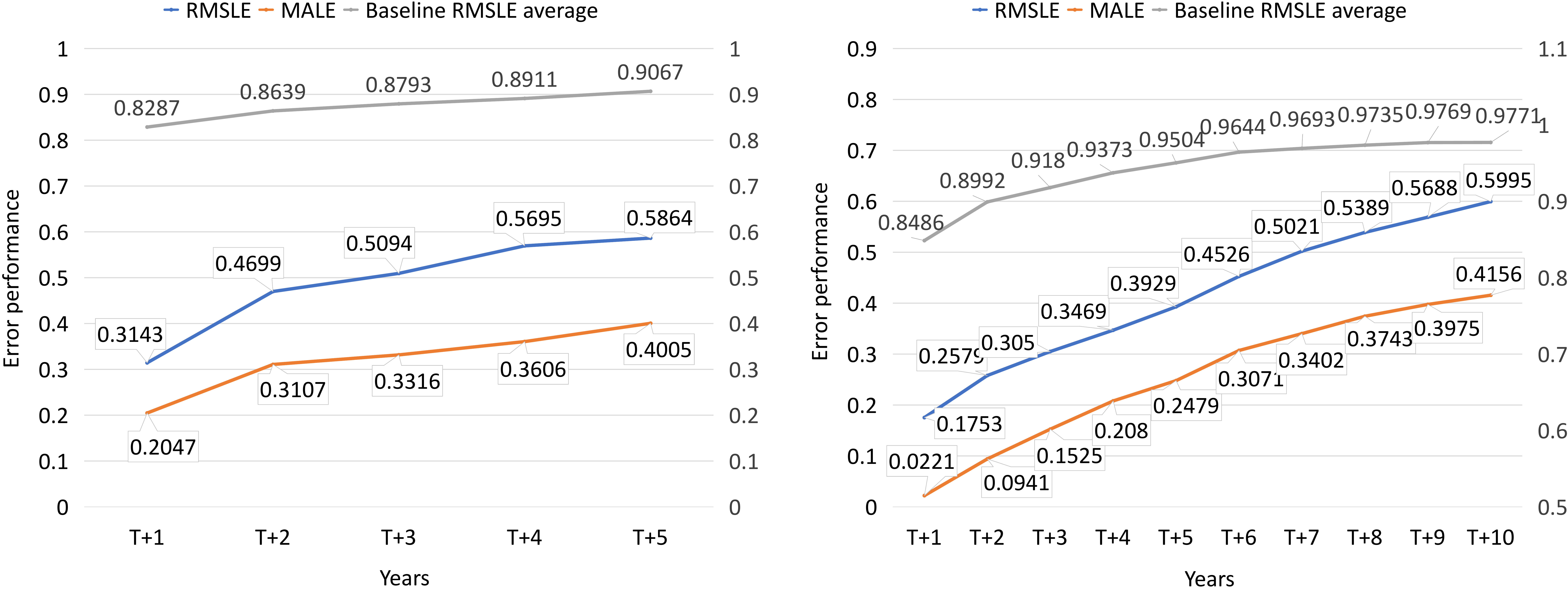}
    \caption{Prediction performance depending on year distances after the current time $T$. CTPIR and baselines are evaluated for 5 years on APS (\textbf{Left}) and for 10 years on AIPatent (\textbf{Right}).}
    \label{future}
\end{figure}

We make further analysis on the performance of CTPIR considering different time distances. For target publications, we use snapshots before time $T$ (including $T$) for feature extraction, to predict their cumulative citation counts in the $N_{th}$ year after $T$, denoted $T+N$. The maximum value of $N$ is 5 for APS and 10 for AIPatent. The results are shown in Figure \ref{future}. The precision gradually declines with distance increases, which indicates that a publication's potential impact is much easier to predict in its near future than in the long term. The long-term prediction is based not only on the current influence of the past behavior, but also on some undiscovered changes in the near future. Meanwhile, CTPIR outperforms the average of baselines in all years, particularly in the near future.

\subsection{Prediction Result Analysis.}
\begin{figure}[htbp]
    \centering
    \includegraphics[width=8.5cm]{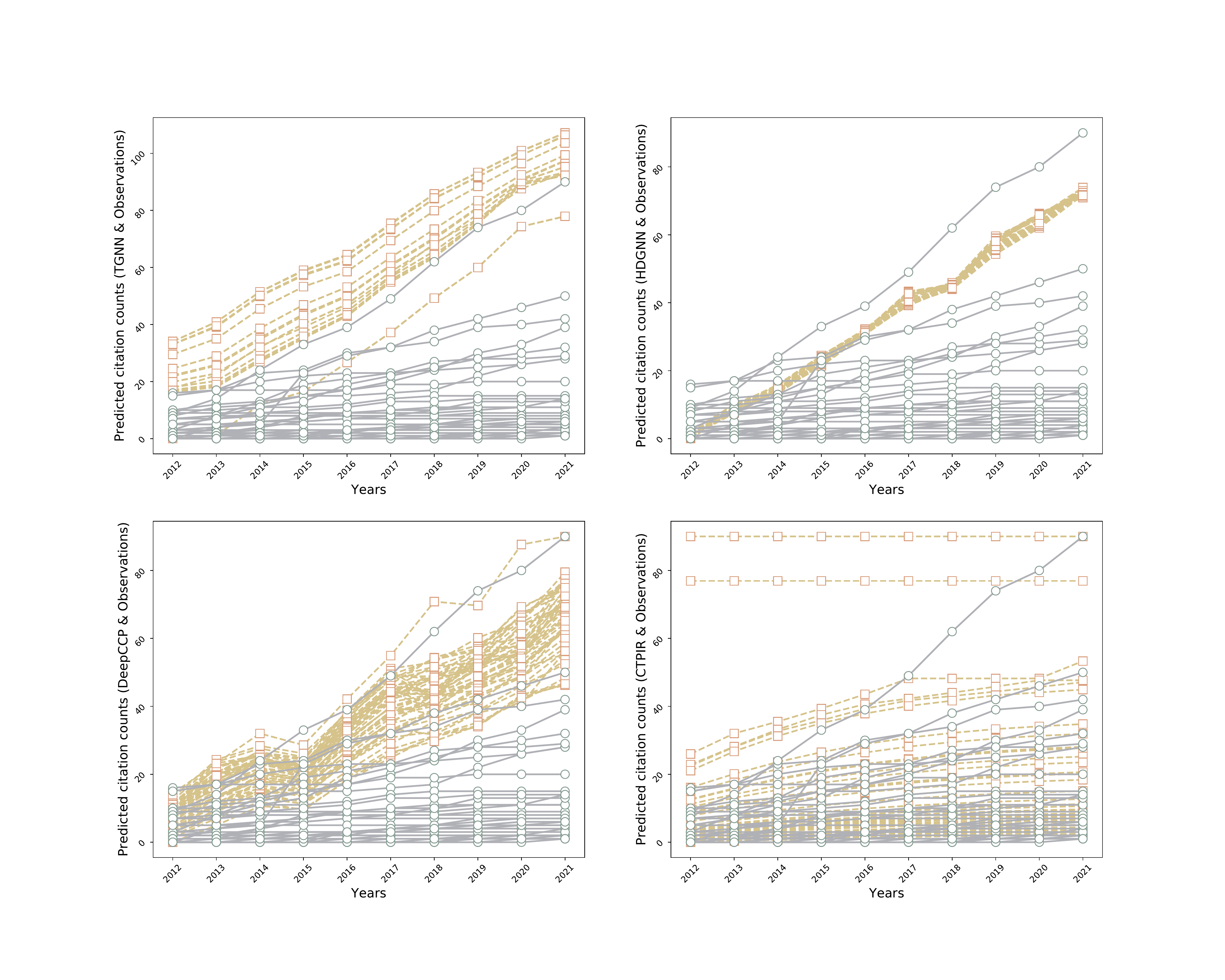}
    \caption{Citation trajectories output from frameworks against the true observations on AIPatent. Above are prediction results of TGNN (\textbf{Left}) and HDGNN (\textbf{Right}). Below are results of DeepCCP (\textbf{Left}) and our CTPIR (\textbf{Right}).}
    \label{prediction}
\end{figure}

To understand why our CTPIR can make such a close prediction to the true observations, we randomly select a number of samples from AIPatent and plot their citation trajectories generated from our proposed framework and some baselines. Figure \ref{prediction} shows that CTPIR can greatly squeeze curves to fit observation lines much better than others. As CTPIR is the only approach to extract temporal features at the attribute level, our proposed influence representation model pays more attention to fine-grained history variations, and finally produces citation numbers closer to real ones. Meanwhile, limiting citation curves to a slower increasing range is helpful to identify outliers in this situation, which leads to social significance in applications. However, some tremendous citation changes cannot be captured well with CTPIR, which needs to be further studied.

\section{Conclusion}
In this paper, we propose \textbf{CTPIR}, a framework for predicting citation trajectory through influence representation using temporal knowledge graphs. Following three motivations listed in Section 4, our proposed framework can represent the influence of publications from a fine-grained perspective with a more expressive temporal knowledge graph learning approach. We also provide a new dataset named \textbf{AIPatent} to facilitate following temporal graph studies. With a comprehensive task design, an improved evaluation strategy and multifaceted analysis are performed to verify the effectiveness of our framework. The ability to capture outlier citation changes can be studied in future work.

\section*{Acknowledgment}
This work is supported by the Key Research and Development Program of Zhejiang Province, China (No. 2021C01013).

\end{document}